\documentclass[10pt,logo,copyright]{nvidiatechreport}
\usepackage{tablefootnote}
\usepackage{array}
\usepackage{cite} 
\usepackage{epsfig}
\usepackage{chngcntr}
\usepackage{amssymb}
\usepackage{amsmath}
\usepackage{graphicx}
\usepackage{verbatim}
\usepackage{float}
\usepackage{booktabs}
\usepackage{multirow}
\usepackage{tabularx}
\usepackage{threeparttable}

\graphicspath{ {./figs/} }

\title{\textbf{COMPASS}: \textbf{C}ross-emb\textbf{O}diment \textbf{M}obility \textbf{P}olicy via Residu\textbf{A}l RL and \textbf{S}kill \textbf{S}ynthesis}

\author[1]{Wei Liu}
\author[1]{Huihua Zhao}
\author[1,2]{Chenran Li}
\author[1]{Yuchen Deng}
\author[1,3]{Joydeep Biswas}
\author[1]{Soha Pouya}
\author[1]{Yan Chang}

\affil[1]{NVIDIA}
\affil[2]{UC Berkeley}
\affil[3]{UT Austin}

\begin{abstract}
As robots are increasingly deployed in diverse application domains, enabling robust mobility across different embodiments has become a critical challenge. Classical mobility stacks, though effective on specific platforms, require extensive per-robot tuning and do not scale easily to new embodiments. Learning-based approaches, such as imitation learning (IL), offer alternatives, but face significant limitations on the need for high-quality demonstrations for each embodiment.

To address these challenges, we introduce COMPASS, a unified framework that enables scalable cross-embodiment mobility using expert demonstrations from only a single embodiment. We first pre-train a mobility policy on a single robot using IL, combining a world model with a policy model. We then apply residual reinforcement learning (RL) to efficiently adapt this policy to diverse embodiments through corrective refinements. Finally, we distill specialist policies into a single generalist policy conditioned on an embodiment embedding vector. This design significantly reduces the burden of collecting data while enabling robust generalization across a wide range of robot designs. Our experiments demonstrate that COMPASS scales effectively across diverse robot platforms while maintaining adaptability to various environment configurations, achieving a generalist policy with a success rate approximately 5X higher than the pre-trained IL policy on unseen embodiments, and further demonstrates zero-shot sim-to-real transfer. 

Project page: \texttt{https://nvlabs.github.io/COMPASS}
\end{abstract}

\begin{document}
\maketitle
\abscontent

\begin{figure*}[h]
\begin{center}
\centering\includegraphics[width=\linewidth]{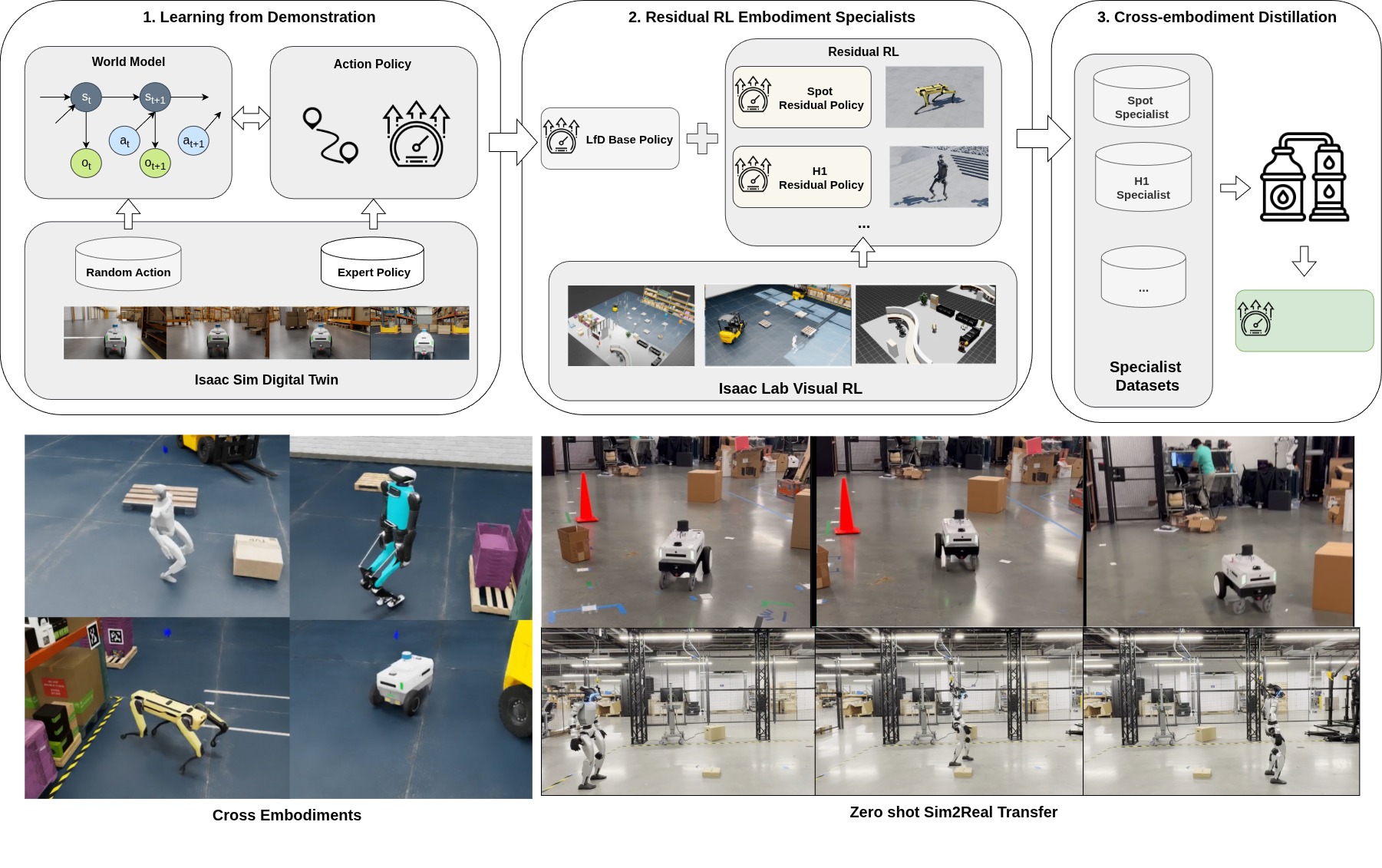}
\caption{High-level overview of the COMPASS workflow: (1) Imitation learning produces a base policy and world model using readily available teacher policies on a mobile robot. (2) Residual RL fine-tunes the base policy for multiple embodiments, optimizing for physical constraints and sensor modalities. (3) Policy distillation consolidates these embodiment-specialist policies into one robust cross-embodiment policy.}
\label{fig:overview}
\end{center}
\end{figure*}

% -------------------------------------------------------------
% 1. Introduction
% -------------------------------------------------------------
\section{Introduction}
\label{sec:intro}

While robotics has made significant strides in both industry and daily life, achieving robust mobility across diverse robot embodiments remains a fundamental open challenge. Morphological differences in hardware, kinematics and sensing configurations \cite{doshi2024scaling, xu2023xskill, yang2024pushing} create large variations in dynamics and perception, complicating efforts to build a universal and adaptable mobility policy for the real world.

Classical mobility stacks \cite{macenski2020marathon2, liu2015autonomous} excel in specific robots, especially wheeled platforms, but often require extensive manual retuning or redevelopment when ported to new embodiments with distinct sensor suites and physical constraints. This reliance on per-robot engineering has motivated interest in end-to-end learning approaches, particularly imitation learning (IL) \cite{liu2024x,xu2021machine}, which aim to scale mobility policies across multiple robots by leveraging data-driven training rather than manual redesign.

However, scaling IL to diverse embodiments introduces a major bottleneck: each new robot typically requires its own set of high-quality demonstrations. As morphological and sensor differences increase, so does the need for additional embodiment-specific data. For complex modalities such as humanoids, collecting demonstrations can be prohibitively expensive or impractical. In parallel, IL can also suffer from the distribution shift \cite{chang2021mitigating}, leading to failures when the policy encounters out-of-distribution states during deployment. While advances in machine learning architectures \cite{wang2023drivedreamer,popov2024mitigating,feng2025survey} and data augmentation techniques \cite{ross2011reduction} help mitigate the distribution shift within a given embodiment, extending these methods to accommodate greater morphological diversity greatly amplifies data requirements and training complexity, further complicating the scalability of pure IL approaches.

Nevertheless, we observe that many key aspects of mobility are inherently shared across robot embodiments, such as environmental understanding, obstacle avoidance, and goal reaching. This observation motivates our central insight: it should be possible to build a cross-embodiment mobility policy by leveraging expert demonstrations from only a single embodiment. 

Based on this principle, we propose COMPASS, a unified three-stage workflow that combines imitation learning, residual reinforcement learning, and policy distillation, as shown in Figure~\ref{fig:overview}. We first pre-train a mobility policy on a single embodiment using IL, leveraging expert demonstrations to train a shared representation of mobility priors over environmental states and robot actions. Following the X-Mobility approach\cite{liu2024x}, this pre-trained model is decomposed into two components: (i) a world model that encodes environmental dynamics, and (ii) a policy that learns to produce actions. To efficiently adapt this policy to various embodiments, we integrate residual reinforcement learning (RL)\cite{johannink2019residual,ankile2024imitation,li2023residual} into our workflow. Rather than training policies from scratch, residual RL enables each embodiment-specific specialist to refine the pre-trained policy through small corrective adjustments, improving closed-loop performance with fewer environment interactions. Finally, we collect data from these specialists and apply policy distillation to merge their expertise into a single cross-embodiment model, conditioned on an embodiment embedding vector. This workflow amortizes the demonstration effort, enabling robust generalization across diverse platforms while requiring expert demonstrations from only a single embodiment. Experiments show that COMPASS achieves a 5X increase in task success rate and a 3X improvement in travel efficiency compared to the IL-only policy, demonstrating both its effectiveness and scalability.

To the best of our knowledge, this three-stage framework is the first end-to-end pipeline for cross-embodiment mobility that systematically addresses morphological diversity, domain shift, and data efficiency. Our key contributions are as follows:
\begin{itemize}
    \item We propose a unified three-stage workflow that combines imitation learning, residual reinforcement learning, and policy distillation to achieve scalable, data-efficient cross-embodiment mobility from demonstrations collected on a single embodiment. 
    \item We develop a visual RL approach that leverages a world model to efficiently refine policies across embodiments by incorporating temporal and spatial structure.
    \item We demonstrate extensive benchmarks across diverse robot platforms, showing that the proposed approach achieves robust generalization while preserving the adaptability needed to succeed in varied environments and zero-shot sim-to-real transfer. 
\end{itemize}

% -------------------------------------------------------------
% 4. Approach / Methodology
% -------------------------------------------------------------
\section{Method}
\label{sec:approach}
We present a three-stage workflow aimed at building robust cross-embodiment mobility policies (see Fig. \ref{fig:overview}). First, we train a base policy via IL, which captures general mobility priors from teacher demonstrations on mobile robots. Next, we refine this base policy into embodiment-specific specialists via residual RL. Finally, policy distillation combines these specialists into a single model suitable for multi-platform deployment.

\begin{figure*}[t]
    \centering
    \includegraphics[width=\linewidth]{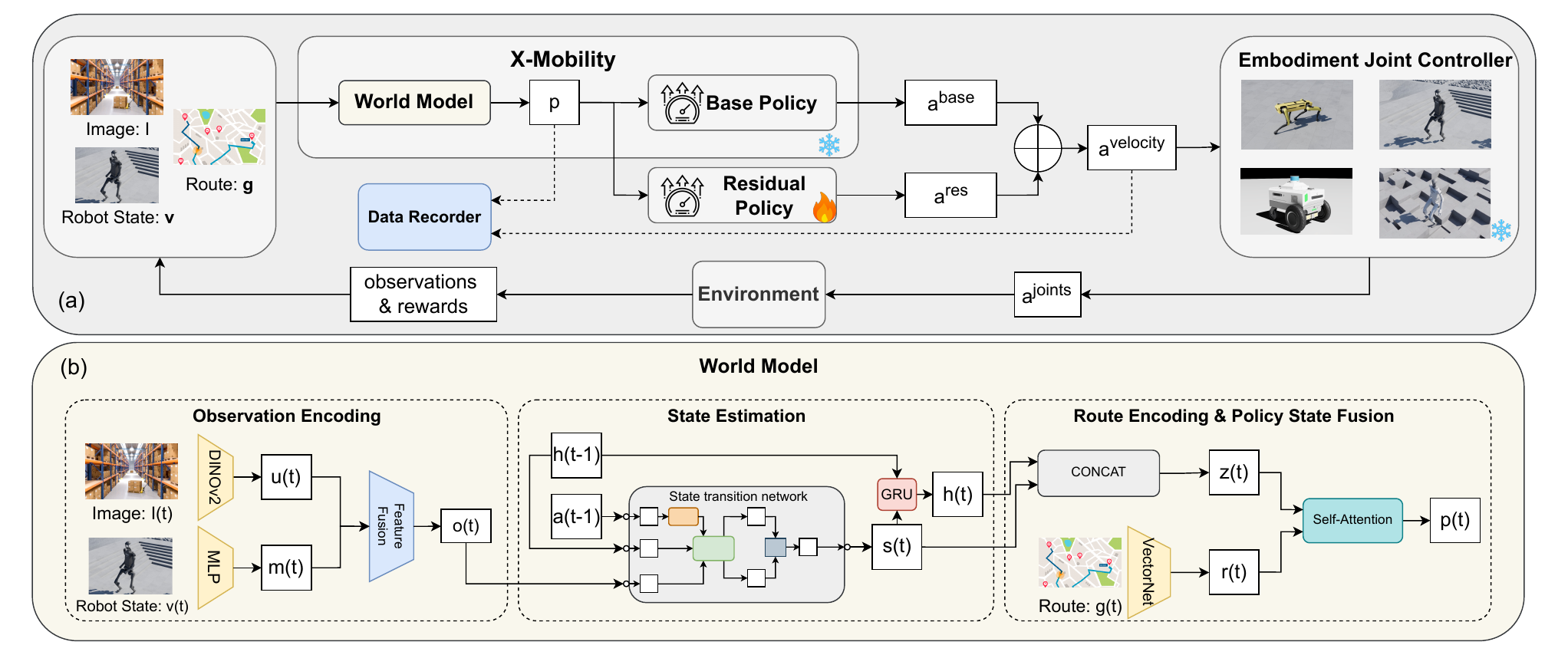}
    \caption{Residual RL architecture: (a) residual RL loop and (b) world model architecture. The world model processes the same inputs as the IL approach to produce the policy state, while the imitation-learned base policy generates a base action. The residual policy refines this action with a correction term, producing the final velocity command for embodiment-specific joint controllers. With the joint actions, the robot interacts with the environment and receives new observations and rewards. The data recorder records the pairs of policy state and action for policy distillation.}
    \label{fig:residual_rl_arch}
\end{figure*}

% -----------------------------------------------------------
\subsection{Problem Statement}
\label{sec:problem}

We focus on the task of point-to-point mobility across different robotic embodiments, each characterized by unique kinematics and dynamics. At time step $t$, let the robot observe a state
\[
    \mathbf{x}_t \;=\; \bigl(\mathbf{I}_t,\, \mathbf{v}_t,\, \mathbf{g}_t,\, \mathbf{e}\bigr),
\]
where $\mathbf{I}_t$ is current camera input (RGB images), $\mathbf{v}_t$ is the measured velocity, $\mathbf{g}_t$ provides route or goal-related information (e.g., goal position in robot frame), and $\mathbf{e}$ is an embodiment embedding that specifies the robot’s morphology. Although $\mathbf{e}$ remains constant for a single robot during an episode, it varies across different embodiments.

We aim to learn a policy $\pi_\theta$ that maps $\mathbf{x}_t$ to a velocity command $\mathbf{a}_t = (v_t, \omega_t)$, which is then consumed by a low-level controller for joint-level actuation. The environment’s transition dynamics $p(\mathbf{x}_{t+1}\mid\mathbf{x}_t,\mathbf{a}_t)$ depend on both the robot’s embodiment and external factors in the scene. We define a reward function $R(\cdot)$ that encourages efficient, collision-free progress to the goal. The objective is to maximize the expected discounted return
\[
    J(\theta) \;=\; \mathbb{E}\left[\sum_{t=0}^{T} \gamma^t\, R(\mathbf{x}_t, \mathbf{a}_t)\right].
\]
The challenge is to design a single policy that leverages the embodiment embedding $\mathbf{e}$, allowing shared knowledge yet accommodating distinct morphological constraints.

% -----------------------------------------------------------
\subsection{Step 1: Imitation Learning for Mobility Priors}
\label{sec:il}

The initial step uses IL to acquire a generic mobility baseline. We rely on readily available teacher policies—often classical mobility stacks—for standard mobile robots, which typically provide reliable demonstrations.

\subsubsection{Latent State Modeling}
We introduce a latent state $\mathbf{s}_t$ to capture environment dynamics. Let $\mathbf{o}_t=\bigl(\mathbf{I}_t,\, \mathbf{v}_t\bigl)$ denote raw observations, including RGB images and robot velocities. Our goal is to learn a world model that predicts transitions in this latent space:
\[
    \mathbf{s}_{t} \;=\; f_{\phi}\bigl(\mathbf{s}_{t-1},\, \mathbf{a}_{t-1}\bigr),
    \quad
    \widehat{\mathbf{o}}_{t} \;=\; g_{\psi}\bigl(\mathbf{s}_t\bigr),
\]

where $f_\phi$ updates the latent state based on the previous action, and $g_\psi$ attempts to reconstruct or predict $\mathbf{o}_t$. Training minimizes reconstruction and predictive losses over a dataset of expert demonstrations
\[
    \mathcal{D} \;=\; \Bigl\{(\mathbf{o}_{0}, \mathbf{a}_{0}^{\star}, \ldots, \mathbf{o}_{T}, \mathbf{a}_{T}^{\star})\Bigr\},
\]
where $\mathbf{a}_t^{\star}$ are the expert (teacher) actions.

\subsubsection{Policy Learning in Latent Space}
Having learned a latent transition, we then train a policy $\pi_{\theta}^{\text{IL}}$ that takes policy state $\mathbf{p}_t$, which is fused from the latent state $\mathbf{s}_t$ and route embedding $\mathbf{r}_t$, to predict $\mathbf{a}_t$:
\begin{align*}
    \mathbf{r}_t &= f_{\theta}(\mathbf{g}_t), \\
    \mathbf{p}_t &= \Phi( \mathbf{s}_t, \mathbf{r}_t), \\
    \mathbf{a}_t &= \pi_{\theta}^{\text{IL}}\bigl(\mathbf{p}_t\bigr),
\end{align*}
where $f_{\theta}$ denotes route encoding and $\Phi$ is a learnable feature fusion block.

We minimize the action discrepancy between the policy’s outputs and the teacher’s actions:
\[
    \min_{\theta} \sum_{t} \ell \Bigl(\pi_{\theta}^{\text{IL}}(\mathbf{p}_t), \mathbf{a}_t^{\star}\Bigr),
\]
where $\ell(\cdot)$ could be a simple regression loss. This yields an IL-based mobility priors. The world model helps the policy generalize to out-of-distribution states by predicting future observations and latent transitions, thereby providing a robust encoded representation for decision-making.

\subsubsection{X-Mobility}
We leverage the X-Mobility~\cite{liu2024x} as our base policy, which integrates an autoregressive world model (see Fig. \ref{fig:residual_rl_arch}) with a velocity-prediction policy model. The learned latent state $\mathbf{s}_t$ encapsulates environment dynamics and constraints, while the policy head combines this state with route information to generate velocity commands. X-Mobility’s strong generalization performance indicates that its learned representation readily adapts to different embodiments.

\subsection{Step 2: Residual RL to Fine-Tune Specialists}
\label{sec:residual_rl}

Having trained a promising general mobility policy through IL, we refine it for embodiment-specific needs via residual RL. This stage addresses robot-specific kinematics, sensor configurations, and other constraints that the base IL policy may not fully capture.

\subsubsection{Residual Policy Setup}
Let $\mathbf{a}^{\text{base}}_t \;=\; \pi_{\theta}^{\text{base}}(\mathbf{p}_t)$ be the action from the IL baseline. We introduce a residual policy $\pi_{\phi}^{\text{res}}$ that takes residual policy state $\hat{\mathbf{p}}_t$ and outputs $\mathbf{a}^{\text{res}}_t$. The final action is
\[
    \mathbf{a}_t \;=\; \pi_{\theta}^{\text{base}}(\mathbf{p}_t) + \pi_{\theta}^{\text{res}}(\hat{\mathbf{p}}_t).
\]
The role of $\pi_{\phi}^{\text{res}}$ is to adapt the base policy to nuances of a specific embodiment’s characteristics.

\subsubsection{Residual Policy State}
To enable effective residual policy learning, we introduce a residual policy state $\hat{\mathbf{p}}_t = \mathcal{G}(\mathbf{p}_t)$, which augments the base policy state $\mathbf{p}_t$ with additional task-relevant information. This design allows the residual policy to complement the base policy by incorporating inputs that were not considered during the base policy training. For example, if the base policy omits goal heading information, we can inject its embedding into the residual policy state, allowing the final composite policy to account for the desired heading and achieve more stable goal-reaching behavior. This modular formulation offers flexibility to extend the capabilities of the base policy without retraining it from scratch.

\subsubsection{Residual Policy Network Architecture}
The residual policy network reuses the same policy model as in the base policy. We initialize it by copying the weights of the base action policy and introducing a linear projection layer to align the dimensionality between the base policy state and the residual policy state. The final output layer is reinitialized to exclusively learn the residual correction. This initialization strategy stabilizes training and encourages the residual component to specifically address embodiment-specific discrepancies. The critic network is implemented as a standard multi-layer perceptron (MLP), taking the same residual policy state as input.

\subsubsection{Reward Design}
We define a reward function $R$ to promote safe and efficient mobility, consisting of the following components:
\begin{itemize}
    \item Progress: Positive reward proportional to the reduction in distance to the goal.
    \item Collision avoidance: Penalties for collisions or fall downs.
    \item Goal completion: Large positive reward on reaching the destination with zero velocities.
\end{itemize}
We adopt this simple formulation to facilitate training, while acknowledging that more sophisticated reward shaping could potentially yield better performance.

\subsubsection{Training Loop}
We employ a PPO-based RL optimizer \cite{schulman2017proximal} for the residual policy $\pi_{\phi}^{\text{res}}$. As illustrated in Fig. \ref{fig:residual_rl_arch}, each training iteration proceeds as follows:
\begin{enumerate}
    \item The agent receives the current state $\mathbf{x}_t$, processes it through the world model to form the policy state $\mathbf{p}_t$, then generates the base action $\mathbf{a}^{\text{base}}_t$ from the IL policy and the residual action $\mathbf{a}^{\text{res}}_t$ from the residual network. 
    \item The combined action $\mathbf{a}_t$ is executed in the simulation environment via an embodiment-specific joint controller.
    \item The agent observes the next state $\mathbf{x}_{t+1}$ and the reward $R$ associated with the transition. 
    \item The residual policy $\pi_{\phi}^{\text{res}}$ is updated via gradient-based methods, while the IL policy $\pi_{\theta}^{\text{base}}$ remains frozen.
\end{enumerate}
The environment resets if the robot reaches the destination, collides with an obstacle, or times out. Upon receiving the reset signal, history states within the world model are also cleared.

By building on a robust pre-trained base policy, the residual RL framework mitigates the sparse sampling challenge. This enables faster convergence to high-performance policies for each embodiment.

% -----------------------------------------------------------
\subsection{Step 3: Policy Distillation to Combine Specialists}
\label{sec:distillation}

After separately training residual RL specialists for each robot embodiment, we consolidate them into a single multi-embodiment policy. This “distilled” policy captures the collective knowledge of all specialist policies while using an embodiment embedding to generalize across different robot platforms.

\begin{figure}[t]
\begin{center}
\centering\includegraphics[width=0.8\linewidth]{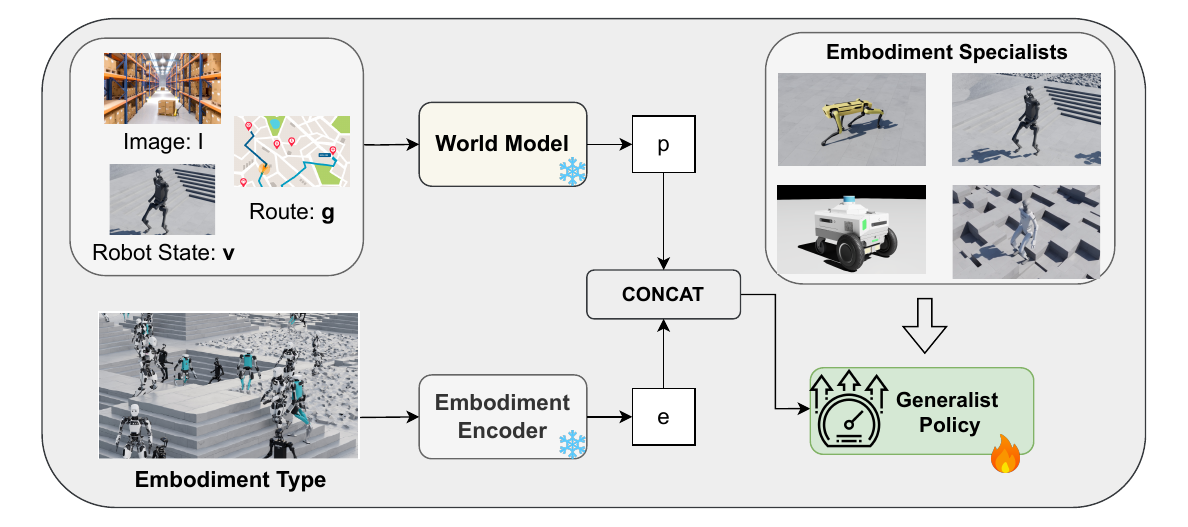}
\caption{Policy distillation aggregates multiple expert specialists (one per embodiment). The final multi-embodiment policy uses a one-hot or learned embedding to condition decisions on robot morphology.}
\label{fig:policy_distillation}
\end{center}
\end{figure}

\subsubsection{Data Collection from Specialists}
After the residual RL training, we record each specialist’s input and output distributions, including:
\begin{itemize}
    \item Policy state from the world model.
    \item One-hot embodiment identifier $\mathbf{e}$.
    \item Mean and variance of the Gaussian action distribution used in PPO.
\end{itemize}
This logged dataset forms the basis for distillation.

\subsubsection{Distillation Method}
Let $\pi_\phi^{(i)}$ denote the specialist policy for the $i$-th embodiment. Each specialist produces a normal distribution $\mathcal{N}(\boldsymbol{\mu}^{(i)}(\mathbf{p}), \sigma^2)$ over actions. We define a distilled policy $\pi_{\theta}^{\text{dist}}$ that outputs $\boldsymbol{\mu}_\theta(\mathbf{p}, \mathbf{e})$ given $\mathbf{p}$ and an embodiment embedding $\mathbf{e}$. To match the specialists’ distributions, we minimize the KL divergence:
\[
    \min_{\theta} \sum_{i \in \mathcal{E}} \sum_{ d^{i}\in \mathcal{D}_i}
    \mathrm{KL}\biggl(
        \mathcal{N}\bigl(\boldsymbol{\mu}^{(i)}(\mathbf{p}), \sigma^2\bigr)
        \,\big\|\, 
        \mathcal{N}\bigl(\boldsymbol{\mu}_\theta(\mathbf{p}, \mathbf{e}^{(i)}), \sigma_\theta^2\bigr)
    \biggr),
\]
where $d^{i}=(\mathbf{p},\, \boldsymbol{\mu}^{(i)}, \sigma^2)$ and $\mathcal{D}^{(i)}$ is the dataset of recorded state-action distributions from the $i$-th specialist. $\mathbf{e}^{(i)}$ is the corresponding embodiment embedding.

\subsubsection{Embodiment Embedding}
A key component of policy distillation is the embodiment embedding $\mathbf{e}$, which captures the morphological and dynamical characteristics of each embodiment. In the simplest version, we use a one-hot encoding vector of length $N$, where $N$ denotes the number of robot embodiments. Each position in this vector corresponds to a specific robot. This straightforward approach is efficient when $N$ is small and the robots differ substantially. We anticipate that a learnable embedding could better generalize to new, unseen embodiments by interpolating within the embedding space, and we leave such zero-shot generalization for future work.

\subsubsection{Distilled Policy Network Architecture}
The distilled policy retains the same latent processing pipeline but additionally conditions on the embodiment embedding before generating the final action distribution (see Fig. \ref{fig:policy_distillation}). The network consists of an MLP for mean prediction and a global variance parameter, resulting in a single policy that achieves near-expert performance across all considered robot types.

Consequently, our three-step framework—imitation learning, residual RL, and policy distillation—bridges the gap between generic mobility knowledge and highly specialized embodiment constraints, yielding a unified cross-embodiment mobility policy.

% -------------------------------------------------------------
% 5. Experimental Setup
% -------------------------------------------------------------
\section{Experimental Setup} \label{sec:exp-setup}

\begin{figure}[t] 
\begin{center} \centering\includegraphics[width=\linewidth]{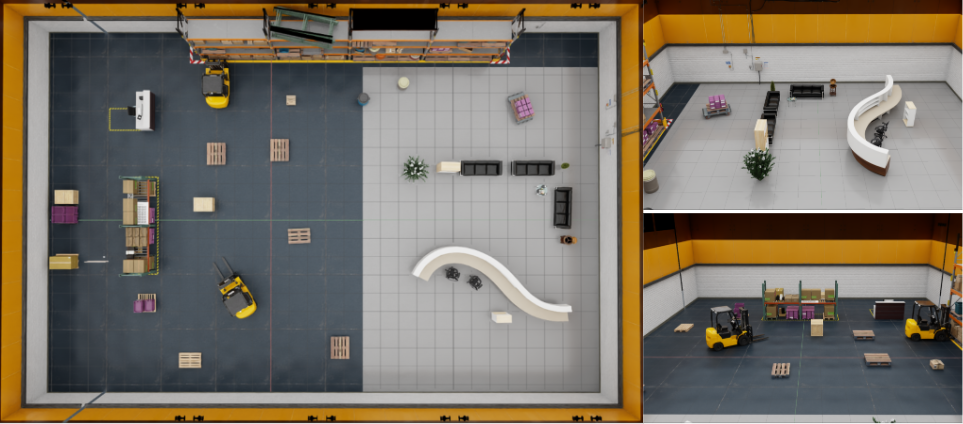} 
\caption{Residual RL training environment. Multiple instances are tiled and run in parallel to accelerate data collection and model updates.} \label{fig:residual_rl_training_env} 
\end{center} 
\end{figure}

\subsection{Training}

\subsubsection{IL Base Policy} For the initial IL stage, we employ a X-Mobility checkpoint pre-trained on Carter dataset. We freeze this checkpoint, which then serves as the base network for subsequent RL refinements.

\subsubsection{Residual RL} 

We utilize Nvidia Isaac Lab\cite{mittal2023orbit} to train our policies in a parallelized visual RL environment, enabling efficient data collection and rapid training updates.

To avoid overfitting and preserve the base policy’s ability to generalize across environments, we construct a diverse set of training scenarios (Fig.~\ref{fig:residual_rl_training_env}) that accommodate four distinct robot embodiments: Nova Carter (wheeled), Unitree H1 (humanoid), Unitree G1 (humanoid), and Spot Mini (quadruped). For the humanoid and quadruped robots, we employ RL-based locomotion policies, trained within Isaac Lab, to map velocity commands to joint-level controls. Due to limited wheel-physics support in Isaac Lab, Nova Carter instead uses a custom controller that directly adjusts the robot’s root state based on velocity commands.

Each embodiment is trained in a unified environment that randomly initializes the agent’s pose and the goal location, where the goal distance is uniformly sampled between 2m and 5m from the robot’s starting position. The straight line between the robot and the goal serves as a simplified route, providing short-horizon guidance within the camera’s field of view. Each episode spans up to 256 timesteps and resets if the agent collides, reaches its goal, or exceeds the maximum episode length. We train each embodiment specialist for 1,000 episodes with 64 environments in parallel using 2 Nvidia L40 GPUs, except for Carter, which is trained for only 300 episodes to mitigate overfitting. This reduced training schedule for Carter is necessary because X-Mobility is already trained on Carter dataset, making it prone to overfitting if extensively fine-tuned.

\subsubsection{Policy Distillation} To distill the learned specialists into a single unified policy, we record 320 trajectories per embodiment using the same environment as residual RL training. Each trajectory spans 128 steps, yielding approximately 40k frames per embodiment. We then perform policy distillation training on 4 Nvidia H100 GPUs by aligning each specialist’s output distribution.

\subsection{Benchmark}

\subsubsection{Metrics} We evaluate both efficiency and safety using two key metrics: 
\begin{itemize} 
    \item Success Rate (SR): Fraction of trials that reach the goal region without collision or timeout. 
    \item Weighted Travel Time (WTT): Total travel time to reach the goal, conditioned on success, divided by SR. 
\end{itemize}

\subsubsection{Scenarios}

\begin{figure}[t] 
\begin{center} \centering\includegraphics[width=0.7\linewidth]{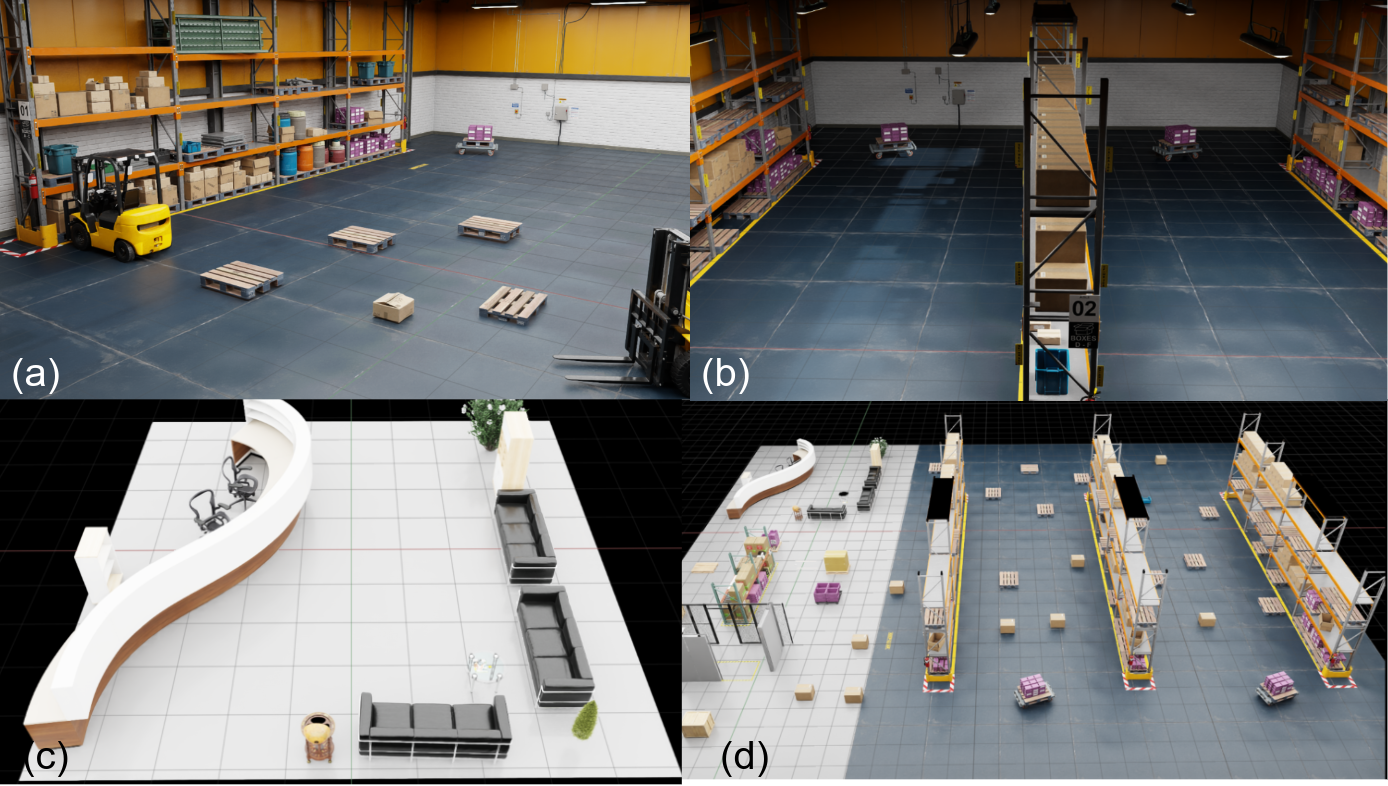} 
\caption{Evaluation environments with multiple layouts to assess policy generalization: (a) warehouse with single rack ($24m\times38 m$), (b) warehouse with multi racks ($24m\times38 m$), (c) office ($10m\times10 m$), and (d) combined scenes with multi racks ($30m\times 38 m$).} 
\label{fig:rl_evaluation_env} 
\end{center} 
\end{figure}

We evaluate our algorithm in four environments of increasing difficulty as in Fig. \ref{fig:rl_evaluation_env}. In the simplest environment, only sparse low lying obstacles are present, while in the complex one, a warehouse with multiple racks requires long-horizon mobility under minimal route guidance. To further assess the policy’s robustness and generalization, we also vary textures and lighting conditions across these environments.

For each environment, we conduct 640 trials with randomly sampled initial and goal poses. Each trial is terminated after 25.6s of execution. We then collect performance metrics for each embodiment and environment type, allowing direct comparison of mobility performance across different settings.

\section{Results} \label{sec:results}

This section addresses four primary questions: 

\begin{itemize}
    \item Does residual RL effectively yield high-performance embodiment specialists from the IL baseline? 
    \item Can policy distillation consolidate these specialists into a single robust generalist without a considerable performance drop? 
    \item How does our approach compare to alternative strategies, such as training RL from scratch? 
    \item Can COMPASS policy trained in simulation be applied to real robot platforms?
\end{itemize}

\subsection{Does residual RL produce high-performance embodiment specialists?} 
To investigate whether residual RL can improve performance relative to the IL baseline (X-Mobility), we compared the specialist policies for each embodiment against X-Mobility, and the results are summarized in Table~\ref{tab:benchmark}. In all evaluations, the specialist policies significantly outperform X-Mobility across both success rate (SR) and mobility efficiency (WTT). The SR is improved with a factor ranging from $5$x to $40$x, and the WTT is improved $3$x on average. This performance boost suggests that residual RL effectively leverages environment semantics and embodiment-specific physics. In particular, by exploring each robot’s physical limits during training, the specialist can produce commands that maximize efficiency while maintaining balance and stability.

A notable observation is that while X-Mobility performs best on the Carter robot, its zero-shot performance on other embodiments degrade significantly. This aligns with expectations and underscores the need for embodiment-specific policy adaptations to ensure robustness. Compared to the original X-Mobility paper, we observe a performance drop for Carter in warehouse environments within Isaac Lab, primarily due to simplified routes and variations in image rendering quality. This motivates future improvements, such as incorporating a hierarchical mobility stack for more sophisticated routing and exposing the model to diverse rendering conditions during latent state training.

\begin{table}[h]
    \centering
    \small
    \caption{Benchmark Results}
    \begin{tabular}{c|c|cc|cc|cc|cc}
        \toprule
        \textbf{Emb.} & \textbf{Model} & \multicolumn{2}{c|}{\textbf{Warehouse Single}} & \multicolumn{2}{c|}{\textbf{Warehouse Multi}} & \multicolumn{2}{c|}{\textbf{Office}} & \multicolumn{2}{c}{\textbf{Combined}} \\
        & & SR(\%) & WTT(s) & SR(\%) & WTT(s) & SR(\%) & WTT(s) & SR (\%) & WTT(s) \\
        \midrule
        \multirow{3}{*}{Carter} 
        & X-Mobility & 42.3 & 8039 & 37.9 & 9351 & 50.6 & 5508 & 50.3 & 5319 \\
        & Carter Specialist & \textbf{91.5} & 2282 & 91.8 & \textbf{2283} & 72.0 & \textbf{2340} & \textbf{85.4} & \textbf{2442} \\
        & Generalist & 90.6 & \textbf{2241} & \textbf{91.9} & 2361 & \textbf{73.4} & 2504 & 85.3 & 2527 \\
        \midrule
        \multirow{3}{*}{H1} 
        & X-Mobility & 17.5 & 12588 & 9.2 & 12234 & 25.6 & 10791 & 22.0 & 11519 \\
        & H1 Specialist & \textbf{94.5} & \textbf{4123} & 87.9 & \textbf{4283} & \textbf{66.7} & 3300 & 82.8 & \textbf{3698} \\
        & Generalist & 93.4 & 4129 & \textbf{88.9} & 4319 & 66.2 & \textbf{3260} & \textbf{84.4} & 3802 \\
        \midrule
        \multirow{3}{*}{Spot} 
        & X-Mobility & 5.7 & 13320 & 4.5 & 12007 & 6.1 & 7740 & 11.7 & 11426 \\
        & Spot Specialist & 84.5 & \textbf{3183} & \textbf{93.5} & \textbf{3204} & \textbf{76.2} & 3587 & 77.1 & 3490 \\
        & Generalist & \textbf{84.7} & 3207 & 93.2 & 3255 & 74.8 & \textbf{3512} & \textbf{77.9} & \textbf{3485} \\
        \midrule
        \multirow{3}{*}{G1} 
        & X-Mobility & 3.6 & 13913 & 1.8 & 13781 & 2.8 & 13276 & 10.0 & 11865 \\
        & G1 Specialist & 95.6 & \textbf{4667} & 93.7 & \textbf{5254} & \textbf{77.0} & \textbf{3960} & 90.0 & \textbf{3968} \\
        & Generalist & \textbf{95.7} & 4717 & \textbf{94.5} & 5360 & 76.7 & 4031 & \textbf{90.6} & 3991 \\
        \bottomrule
    \end{tabular}
    \label{tab:benchmark}
\end{table}

\subsection{Can policy distillation produce a robust generalist?} 
We next evaluate whether policy distillation can merge the specialists into a single generalist policy without sacrificing overall performance. As shown in Table~\ref{tab:benchmark}, the generalist achieves performance comparable to, and sometimes exceeding, that of the specialist policies. This is especially evident on the G1 robot, likely due to increased diversity in training scenarios and embodiments during distillation. These results confirm that our distillation approach is effective and that building a robust generalist from multiple specialists is both feasible and efficient.

\begin{figure}[t] 
\begin{center} \centering\includegraphics[width=0.8\linewidth]{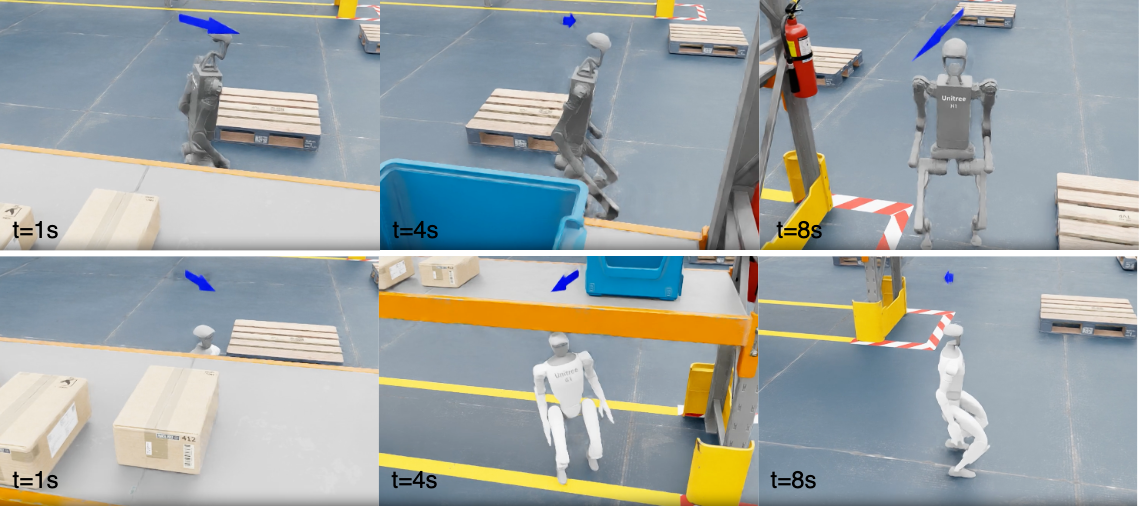} 
\caption{Example illustrates the generalist's cross-embodiment capability: G1 shortcuts beneath the shelf, while H1 must detour due to insufficient height clearance.} 
\label{fig:ce_example} 
\end{center} 
\end{figure}

As a case study of cross-embodiment capabilities, we designed a long-horizon mobility scenario in a multi-rack warehouse. As illustrated in Fig. \ref{fig:ce_example}, G1 can lean slightly to the right and pass under a shelf with just enough head clearance, whereas H1 lacks sufficient clearance and must take a detour. This example highlights how the generalist effectively accounts for each embodiment’s distinct morphological constraints.

\subsection{How does COMPASS compare to training from scratch?}
\begin{figure}[h]
\begin{center} \centering\includegraphics[width=\linewidth]{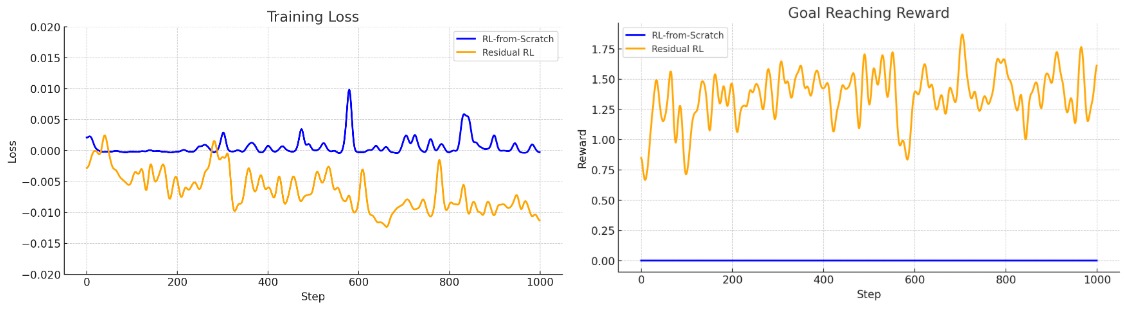} 
\caption{RL from scratch v.s. Residual RL.} 
\label{fig:rl_from_scratch} 
\end{center} 
\end{figure}

We also explored whether we could train a policy from scratch using the same RL setup with the latent state but without leveraging the IL base actions. We observed that this RL-from-scratch approach struggled to converge even after 1,000 episodes as in Fig. \ref{fig:rl_from_scratch}, indicating significantly lower training efficiency. In contrast, our residual RL method exhibited notably faster convergence, underscoring how leveraging an IL baseline resolves the sparsity issues that commonly plague RL in mobility tasks.

\subsection{Can COMPASS Policy trained in simulation be applied to real robot platforms?}
To evaluate the Sim2Real transferability, we deployed it on two real-world robotic platforms powered by the NVIDIA Jetson Orin: Carter and Unitree G1 (Fig.~\ref{fig:overview}). Leveraging the world model latent state representation, along with Isaac Lab’s high-quality visual rendering and accurate physics simulation, both robots achieved robust zero-shot transfer, successfully navigating cluttered environments without any additional fine-tuning. With localization enabled by cuVSLAM\cite{korovko2025cuvslam}, the policy attained a success rate of approximately 80\% across 20 trials with randomly placed obstacles, consistently avoiding collisions and reaching the designated targets. We anticipate that RL fine-tuning focused on semantic understanding of the environment, particularly when combined with a Real2Sim pipeline \cite{zhu2025vr}, can further narrow the Sim2Real gap. 

Moreover, onboard inference with TensorRT requires only $\sim$30 ms per step as in Table \ref{tab:trt_run_time}, demonstrating that the policy is computationally efficient and well-suited for real-time deployment.

\begin{table}[h]
\centering
\caption{Inference Latency and GPU Memory Usage}
\begin{tabular}{c c c c c}
\toprule
\textbf{Model} & \textbf{P50(ms)} & \textbf{P95(ms)} & \textbf{GPU Memory(MB)} \\ 
\midrule
\textbf{Distilled Generalist} & 29.3389 & 29.3767 & 422 \\ 
\bottomrule
\end{tabular}
\label{tab:trt_run_time}
\end{table}

\section{Ablation Studies} \label{sec:ablation}

\subsection{Residual RL}
In this section, we present ablation studies evaluating the various design decisions in our residual RL approach. We exclude the Carter robot from these experiments as its cross-embodiment effect is less pronounced.

\begin{table}[h]
    \centering
    \small
    \caption{Ablation Studies on Residual RL.}
    \begin{tabular}{c|c|cc|cc|cc|cc}
        \toprule
        \textbf{Emb.} & \textbf{Model} & \multicolumn{2}{c|}{\textbf{Warehouse Single}} & \multicolumn{2}{c|}{\textbf{Warehouse Multi}} & \multicolumn{2}{c|}{\textbf{Office}} & \multicolumn{2}{c}{\textbf{Combined }} \\
        & & SR(\%) & WTT(s) & SR(\%) & WTT(s) & SR(\%) & WTT(s) & SR (\%) & WTT(s) \\
        \midrule
        \multirow{5}{*}{H1} 
        & \textbf{Default} & 94.5 & 4123 & 87.9 & 4283 & 66.7 & 3300 & \textbf{82.8} & 3698 \\
        & w/ Curriculum & \textcolor{ForestGreen}{96.1} & 4045 & \textcolor{ForestGreen}{88.7} & 4155 & \textcolor{red}{66.5} & 3148 & \textcolor{ForestGreen}{83.9} & 3933 \\
        & Depth Critic & \textcolor{red}{90.9} & 4245 & \textcolor{red}{87} & 4254 & \textcolor{ForestGreen}{72.3} & 3530 & \textcolor{ForestGreen}{85.4} & 3764 \\
        & Warehouse Only & \textcolor{ForestGreen}{94.6} & 3237 & \textcolor{ForestGreen}{88.7} & 3249 & \textcolor{red}{64.0} & 2850 & \textcolor{red}{81.0} & 3152 \\
        \midrule
        \multirow{4}{*}{Spot} 
        & \textbf{Default} & 84.5 & 3183 & 93.5 & 3204 & 76.2 & 3587 & 77.1 & 3490 \\
        & w/ Curriculum & \textcolor{red}{81.5} & 3079 & \textcolor{red}{86.2} & 3128 & \textcolor{red}{66.1} & 3112 & \textcolor{red}{71.5} & 3238 \\
        & Depth Critic &\textcolor{red}{83.4} & 3090 & \textcolor{red}{92.1} & 3125 & \textcolor{ForestGreen}{77.9} & 3645 & \textcolor{ForestGreen}{77.6} & 3226 \\
        \midrule
        \multirow{4}{*}{G1} 
        & \textbf{Default} & 95.6 & 4667 & \textbf{93.7} & 5254 & \textbf{77.0} & 3960 & \textbf{90.0} & 3968  \\
        & w/ Curriculum & \textcolor{ForestGreen}{96.4} & 4376 & \textcolor{red}{86.5} & 4992 & \textcolor{red}{69.6} & 3478 & \textcolor{red}{89.0} & 4018 \\
        & Depth Critic & \textcolor{ForestGreen}{95.7} & 4600 & \textcolor{red}{87.9} & 5153 & \textcolor{ForestGreen}{78.4} & 4091 & \textcolor{ForestGreen}{91.4} & 4018 \\
        \bottomrule
    \end{tabular}
    \begin{tablenotes}
    \item \textcolor{ForestGreen}{Green}/\textcolor{red}{Red}: Better/Worse than default setting
    \end{tablenotes}
    \label{tab:ablation_residual_rl}
\end{table}

\subsubsection{Curriculum on Goal Distance} 
In our RL training setup, we currently do not use a curriculum. To investigate whether such a strategy could improve training efficiency, we introduced a curriculum based on goal distance. Rather than enforcing a minimum goal distance from the outset, we gradually increased the probability of applying this constraint as training progressed to enable smoother convergence.

However, the results in Table \ref{tab:ablation_residual_rl} indicate that the curriculum-based approach did not outperform the baseline without a curriculum. A likely explanation is that the IL base policy already provides a strong initialization, and reducing exposure to challenging scenarios with the curriculum may have limited the policy’s capacity to explore and succeed on more difficult tasks.

\subsubsection{Critic State}
In our default setup, the critic network uses the same policy state as the actor, under the assumption that the policy state already encapsulates sufficient information to learn accurate value functions. While this approach often works well, it depends on implicit information, which may make learning the critic network more challenging. To investigate potential improvements, we explored different configurations for the critic's input. In one variant, we provided a depth image explicitly to the value function by employing a ResNet18 for feature extraction, then concatenating the resulting features with velocity and route embeddings to form the critic state. As shown in Table \ref{tab:ablation_residual_rl}, this strategy enhanced performance in office scenarios—where tall obstacles exist—but inadvertently reduced performance in warehouse environments presented with more low-lying obstacles. These findings suggest that the additional signal from the depth image can boost performance under certain conditions, indicating a promising direction for further exploration.

\subsubsection{Training Environment}
In another ablation study, we evaluate the impact of the training environment. Our default setup uses a combined environment consisting of warehouse, office, and lab scenarios. To assess the effect of environmental diversity, we also conducted a separate training run using only the warehouse environment with H1 robot. The results, shown in Table \ref{tab:ablation_residual_rl}, indicate that restricting training to warehouse scenarios slightly enhances performance in warehouse scenes but reduces performance in the office and combined settings. These findings underscore the value of diverse training environments for improving generalization. Notably, even though the office setting was excluded from this specialized training, the model still performs reasonably well when tested there, demonstrating a degree of environment generalizability.

\begin{table}[t]
    \centering
    \small
    \caption{Ablation study on policy distillation}
    \begin{tabular}{c|c|cc|cc|cc|cc}
        \toprule
        \textbf{ Emb.} & \textbf{Koss / Dataset} & \multicolumn{2}{c|}{\textbf{Warehouse Single}} & \multicolumn{2}{c|}{\textbf{Warehouse Multi}} & \multicolumn{2}{c|}{\textbf{Office}} & \multicolumn{2}{c}{\textbf{Combined}} \\
        & & SR(\%) & WTT(s) & SR(\%) & WTT(s) & SR(\%) & WTT(s) & SR (\%) & WTT(s) \\
        \midrule
        \multirow{3}{*}{Carter} & KL / All & 90.6 & 2241 & 91.9 & 2361 & 73.4 & 2504 & 85.3 & 2527 \\
        & MSE / All & \textcolor{red}{89.8} & 2201 & \textcolor{ForestGreen}{92.1} & 2331 & \textcolor{ForestGreen}{73.7} & 2585 & \textcolor{red}{85.1} & 2452 \\
        & KL / Good & \textcolor{ForestGreen}{93.1} & 2246 & \textcolor{ForestGreen}{92.0} & 2388 & \textcolor{red}{72.3} & 2427 & \textcolor{red}{83.1} & 2744 \\
        \midrule
        \multirow{3}{*}{H1} & KL / All & 93.4 & 4129 & 88.9 & 4319 & 66.2 & 3260 & 84.4 & 3802 \\
        & MSE / All & \textcolor{red}{93.3} & 4082 & \textcolor{red}{87.6} & 4260 & \textcolor{ForestGreen}{66.5} & 3251 & \textcolor{red}{82.1} & 3717 \\
        & KL / Good & \textcolor{ForestGreen}{93.6} & 4043 & \textcolor{red}{88.4} & 4218 & \textcolor{red}{66.1} & 3288 & \textcolor{ForestGreen}{88.4} & 4218 \\
        \midrule
        \multirow{3}{*}{Spot} & KL / All & 84.7 & 3207 & 93.2 & 3255 & 74.8 & 3512 & 77.9 & 3485 \\
        & MSE / All & \textcolor{red}{84.3} & 3180 & \textcolor{ForestGreen}{93.6} & 3218 & \textcolor{ForestGreen}{75.0} & 3512 & \textcolor{ForestGreen}{79.2} & 3516 \\
        & KL / Good & \textcolor{red}{84.6} & 3234 & \textcolor{ForestGreen}{94.3} & 3200 & \textcolor{ForestGreen}{75.1} & 3462 & \textcolor{ForestGreen}{78.4} & 3410 \\
        \midrule
        \multirow{3}{*}{G1} & KL / All & 95.7 & 4717 & 94.5 & 5360 & 76.7 & 4031 & 90.6 & 3991 \\
        & MSE / All & \textcolor{ForestGreen}{96.8} & 4596 & \textcolor{red}{94.3} & 5336 & \textcolor{ForestGreen}{77.6} & 4030 & \textcolor{red}{89.8} & 3934 \\
        & KL / Good & \textcolor{red}{95.1} & 4622 & \textcolor{red}{93.4} & 5180 & \textcolor{red}{75.4} & 3884 & \textcolor{red}{90.0} & 3946 \\
        \bottomrule
    \end{tabular}
    \begin{tablenotes}
    \item \textcolor{ForestGreen}{Green}/\textcolor{red}{Red}: Better/Worse than baseline
    \end{tablenotes}
    \label{tab:ablation_policy_distillation}
\end{table}

\subsection{Policy Distillation}
In our final ablation study, we want to examine the policy distillation process. The default setup trains the distilled policy to minimize KL divergence on an unfiltered dataset that may include failure cases. Our first goal is to assess whether capturing the full policy distribution is more beneficial than merely imitating the mean. To this end, we trained another model using only an MSE loss on the mean predictions. As shown in Table \ref{tab:ablation_policy_distillation}, this approach led to slightly inferior performance, particularly on the H1 robot, suggesting that KL divergence—which preserves variance—provides more flexibility for managing training noise and architectural differences.

We also investigated the role of dataset quality by constructing a new dataset from the same scenarios but excluding all failure cases, hypothesizing that this might yield better overall performance. Contrary to our expectations, the results in Table \ref{tab:ablation_policy_distillation} improved only marginally. We suspect that removing failure cases also eliminates corner cases that could otherwise offer valuable learning opportunities for the generalist policy.

\section{Applications}
This section presents selected applications of the COMPASS workflow and its trained policy, demonstrating capabilities that extend beyond the COMPASS workflow itself.
\subsection{Open Vocabulary Object Navigation}
\begin{figure}[t]
\begin{center} \centering\includegraphics[width=\linewidth]{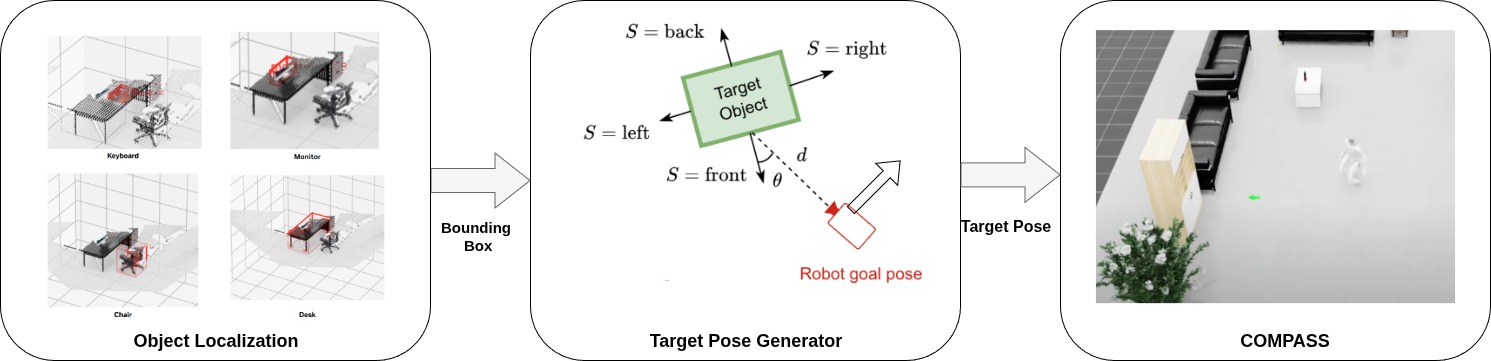} 
\caption{Open vocabulary object navigation by integrating COMPASS and object localization.} 
\label{fig:object_nav} 
\end{center}
\end{figure}

The COMPASS policy is trained for point-based navigation, where the navigation command specifies a target pose. Thanks to its flexible interface design, COMPASS can also support open-vocabulary object navigation by integrating it with an object localization model such as Locate3D \cite{arnaud2025locate} (see Fig. \ref{fig:object_nav}). Given a language prompt describing the target object, the object localization model predicts the corresponding 3D bounding box. Combined with user-defined parameters, such as relative heading and clearance distance, the target pose generator computes an appropriate goal pose for COMPASS to navigate toward. This setup enables a modular framework that extends COMPASS from point-based navigation to open-vocabulary object navigation.

\subsection{Synthetic Datasets for VLA Fine Tuning}
Given the high quality of the RL specialist policy, it can also be leveraged to generate large-scale synthetic datasets in Isaac Lab for training foundational Vision-Language-Action (VLA) models for navigation tasks that typically demand substantial amounts of diverse training data. As a demonstration, we post-trained the Gr00t N1.5 model \cite{bjorck2025gr00t} using the COMPASS distillation dataset generated by the G1 expert policy. In this setup, goal information and measured velocity were provided as state inputs, while ego-centric video frames served as the visual input, paired with a generic text prompt (“Robot Navigation Task”). The Gr00t model was then trained to predict the corresponding velocity commands.

\begin{figure}[t]
\begin{center} \centering\includegraphics[width=0.8\linewidth]{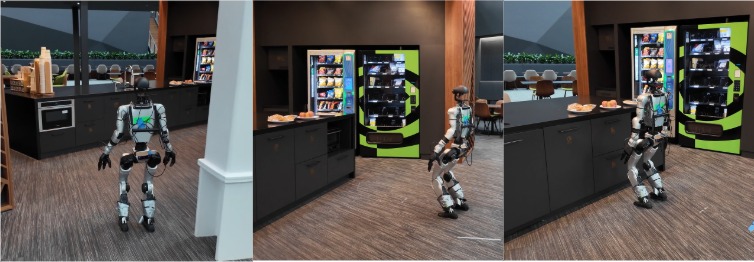} 
\caption{Zero-shot Sim2Real deployment of the post-trained Gr00t model for loco-manipluation tasks.} 
\label{fig:val_sim2real} 
\end{center} 
\end{figure}

Notably, despite being trained on only 600 trajectories and without any prior pre-training on navigation, the fine-tuned Gr00t model successfully acquired robust navigation capabilities. When evaluated on 640 randomly generated test cases in Isaac Lab, the post-trained policy achieved task success rates comparable to the original COMPASS policy in in-distribution warehouse environments, as shown in Table \ref{tab:gn_benchmark}. Moreover, it significantly outperformed COMPASS in the hospital environment, which represents a fully out-of-distribution setting. The model also demonstrated seamless transfer to real-world environments without any additional fine-tuning, enabling smooth integration with the Gr00t manipulation policy for more advanced loco-manipulation tasks, as illustrated in Fig. \ref{fig:val_sim2real}.

\begin{table}[h]
    \centering
    \small
    \caption{Success Rate of Gr00t N1.5 Post-Trained on the COMPASS Dataset}
    \begin{tabular}{c c c}
        \toprule
        \textbf{Model} & \textbf{Warehouse} & {\textbf{Hospital}} \\
        \midrule
        \textbf{Gr00t N1.5} & 86.1\% & 77.6\% \\
        \textbf{COMPASS} & 84.7\% & 45.6\% \\
        \bottomrule
    \end{tabular}
    \label{tab:gn_benchmark}
\end{table}

\section{Future Works}
We have introduced a novel workflow, COMPASS, for cross-embodiment mobility that leverages imitation learning, residual RL, and policy distillation. The initial IL step provides a strong baseline, while residual RL quickly adapts the model to each embodiment's dynamics, and a final distillation step merges specialized policies into a single, generalist policy. Our experiments show that this approach scales effectively to various robot platforms and environments.

Our findings also suggest several key takeaways and directions for future work:
\begin{itemize}
    \item Embodiment Encoding Strategy: We find that one-hot encoding is sufficient when only a few distinct platforms are considered. For a larger or more continuous spectrum of embodiments, a learned embedding might offer richer generalization \cite{johannemann2021sufficientrepresentationscategoricalvariables}.
    \item Distillation Trade-offs: Although policy distillation successfully merges expertise from specialized policies, it can also lead to an averaging effect across embodiments. This phenomenon contributes to embodiment-dependent performance variations.
    \item Transferability vs. Specialization: While residual RL can adapt rapidly to new embodiments, using the same set of hyperparameters can lead to performance divergence, as demonstrated in our studies. Consequently, specialized designs and tuned hyperparameters are often required to achieve optimal results.
    \item Hierarchical Mobility Stack: We observe a decreased success rate in environments such as offices and multi-rack warehouses, primarily for tasks requiring long-horizon route planning. This underscores the importance of a hierarchical mobility stack with a graph-based route planner to handle more complex mobility scenarios.
\end{itemize}

\section{Acknowledgment}
We thank Di Zeng, Chirag Majithia, Sameer Chavan, Billy Okal, Dennis Da, Fernando Castaneda Garcia-Rozas, Peter Varvak, Manuel Chavez Orozco, Vishal Kulkarni, Qi Wang, Yuke Zhu, and Jim Fan for their assistance with system integration and robot testing and for providing insightful feedback on the workflow setup.

\bibliographystyle{IEEEtran}
\bibliography{reference}
\end{document}